\documentclass[letterpaper, 10 pt, conference]{ieeeconf}  

\IEEEoverridecommandlockouts                              

\usepackage{graphics} 
\usepackage{epsfig}
\usepackage{amsmath}
\usepackage{xcolor}
\usepackage{amsfonts}

\usepackage{color}
\usepackage[ruled, vlined]{algorithm2e}

\DeclareMathOperator*{\argmax}{argmax} 
\DeclareMathOperator*{\argmin}{argmin}

\let\labelindent\relax

\usepackage{enumitem}

\title{\LARGE \bf
PrePARE: Predictive Proprioception for Agile Failure Event Detection in Robotic Exploration of Extreme Terrains
}

\author{Sharmita Dey$^{1,2}$, David Fan$^{1,3}$, Robin Schmid${^{1,4}}$, Anushri Dixit$^{5}$, Kyohei Otsu $^{1}$, Thomas Touma$^{6}$, \\
Arndt F. Schilling$^{2}$ 
and Ali-akbar Agha-mohammadi$^{1}$
\thanks{$^{1}$Sharmita Dey, David Fan, Kyohei Otsu and Ali-akbar Agha-mohammadi are with NASA Jet Propulsion Laboratory, California Institute of Technology, Pasadena, CA, USA {\tt\small Sharmita.dey@jpl.nasa.gov}}%
\thanks{$^{2}$Sharmita Dey and Arndt F. Schilling are with the University of Goettingen, University Medical Center Goettingen, Germany}%
\thanks{$^{3}$David Fan is with the Institute for Robotics and Intelligent Machines, Georgia Institute of Technology, Atlanta, GA, USA}%
\thanks{$^{4}$Robin Schmid is with Swiss Federal Institute of Technology (ETH Zürich), 8092 Zurich, Switzerland {\tt\small schmirob@ethz.ch}}%
\thanks{$^{5}$Anushri Dixit is with Control and Dynamical Systems, California Institute of Technology, Pasadena, CA, USA {\tt\small adixit@caltech.edu}}
\thanks{$^{6}$Thomas Touma is with Mechanical and Civil Engineering, California Institute of Technology, Pasadena, CA, USA {\tt\small ttouma@caltech.edu}}%
}

\newcommand{\ph}[1]{{\textbf{#1:}}}

 \newcommand{\textapprox}{\raisebox{0.5ex}{\texttildelow}}

\begin{document}

\maketitle
\thispagestyle{empty}
\pagestyle{empty}

\begin{abstract}
Legged robots can traverse a wide variety of terrains, some of which may be challenging for wheeled robots, such as stairs or highly uneven surfaces. However, quadruped robots face stability challenges on slippery surfaces. This can be resolved by adjusting the robot's locomotion by switching to more conservative and stable locomotion modes, such as crawl mode (where three feet are in contact with the ground always) or amble mode (where one foot touches down at a time) to prevent potential falls. To tackle these challenges, we propose an approach to learn a model from past robot experience for predictive detection of potential failures. Accordingly, we trigger gait switching merely based on proprioceptive sensory information. To learn this predictive model, we propose a semi-supervised process for detecting and annotating ground truth slip events in two stages: We first detect abnormal occurrences in the time series sequences of the gait data using an unsupervised anomaly detector, and then, the anomalies are verified with expert human knowledge in a replay simulation to assert the event of a slip. These annotated slip events are then used as ground truth examples to train an ensemble decision learner for predicting slip probabilities across terrains for traversability. We analyze our model on data recorded by a legged robot on multiple sites with slippery terrain. We demonstrate that a potential slip event can be predicted up to 720 ms ahead of a potential fall with an average precision greater than 0.95 and an average F-score of 0.82. Finally, we validate our approach in real-time by deploying it on a legged robot and switching its gait mode based on slip event detection. 


\end{abstract}

\section{INTRODUCTION}

\begin{figure}[!ht]
    \centering
    \includegraphics[width =\linewidth]{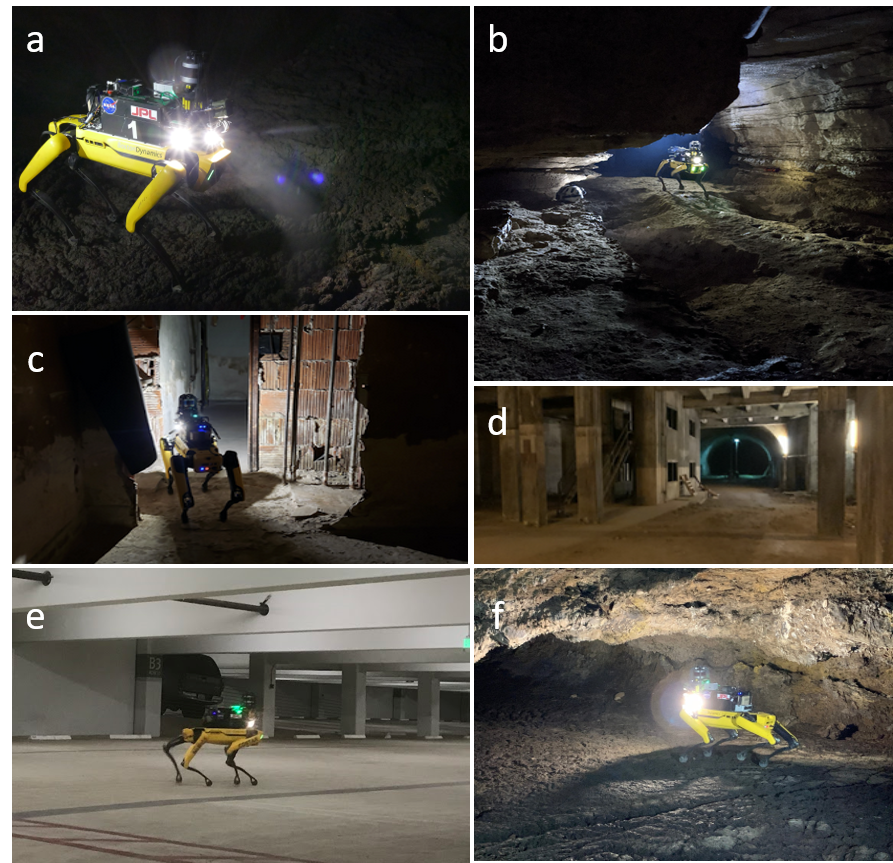}
    \caption{Spot field data collection from different sites: a) and f) Valentine Cave in Lava bed with rough ground surface, steep uniform-slopes and boulders, b) Wells cave, a limestone cave with a slippery surface and steep slopes, c) and d) Abandoned LA subway terminal with dusty terrain, cluttered and narrow doorways) e) LA subway with wide spaces and feature-less surfaces}
    \label{fig:spot_diff_sites}
\end{figure}

Consider a robot navigating in a limestone mine. The robot is on a rescue mission and needs to traverse autonomously in the perceptually degraded environment to report survivors. Traversability in such environments offers several challenges since the robot-ground interactions are particularly difficult to model due to low friction, unstructured inclines, rough terrain, rubble, narrow passages, etc. The robot is vulnerable to slips, imbalance, and falls. If such a situation arises, this may require the robot to adopt recovery maneuvers that would consume time, or even in the worst-case compel the robot to abort the mission if the damage due to fall is beyond recovery. To deal with such cases, a proactive strategy is more appropriate, i.e., it would be particularly helpful if such an edge-case event can be predicted early enough to prepare the robot to prevent a mishap by switching to safer gait modes, e.g., a lower speed gait or a more stable gait mode.  

\ph{Agile adaptation for ground traversability}
Legged robots have the potential to traverse challenging terrains and highly uneven surfaces. 
However, traversing such challenging terrains requires agile adaptation and knowledge of the terrain properties \cite{peng2020learning}. Traversability in such environments poses uncertainty and risk due to high vulnerability to slippage, imbalance, and fall. Slip occurrences are highly non-linear and difficult to model because it happens fast and therefore requires advanced detection. 
Exteroceptive perception such as vision and lidar-based sensing allow looking ahead of the robot and provide a semantic understanding of the terrain in advance to adapt to the terrain properties. 
However, their capabilities of accurate sensing are limited in perceptually degraded environments due to low-light, fog, sparsity, and occlusions. Furthermore, they do not assess the interaction of the robot with the environment like when traversing vegetation, mud or snow. Additionally, acquiring and processing these sensor data is memory-wise and computationally intensive for real-time applications. In contrast, proprioceptive sensing such as joint encoder information-based sensing are robust to these challenges of perceptually difficult environments. They can directly measure the interaction of the robot with the environment, generate data that are memory-wise and computationally less-intensive and are hence more suitable for real-time applications. The main limitation of proprioceptive sensing is the inability to look ahead of the robot. Since proactive adaptation is best to be decided at a fairly premature stage to trigger prevention modules, predictive strategies that can model the robot-environment interaction in advance are suitable for failure event detection like slippage, imbalance or fall.

\ph{Related work} Several works have been done to detect slip and fall during the traversability of autonomous agents. 
A simple way to detect slips is to measure the terrain properties and slip directly using sensors attached to robot end-effectors. In \cite{melchiorri2000slip}, a force/torque sensor attached to the foot-tip of the robot is used to measure the friction coefficient from the contact forces. In \cite{Anandan2015CapacitiveTS} and \cite{Teshigawara2009HighSS}, tactile sensors are proposed which can measure contact forces to detect slips. However, these sensors are prone to damage due to repetitive impacts with the ground when used on legged robots. In \cite{Kajita2004BipedWO}, authors use onboard sensors such as IMU and define a threshold based on acceleration to detect slip events. Further, many of the works depend on classical threshold \cite{focchi2018slip} and state-machine-based approaches \cite{focchi2020heuristic, jenelten2019dynamic, Camurri2017ProbabilisticCE, Bledt2018ContactMF}. An obvious problem of a classical non-learning-based approach is that such heuristics are typically deterministic and  are restricted in extrapolating its functioning to related unseen environments. They are limited to functioning within the predefined scenario and neglect unknown factors such as snow or mud. In contrast, learning-based approaches have the potential to offer generalization to new related scenarios. 
Some learning-based works have incorporated exteroceptive sensor data with proprioceptive data such as \cite{hwangbo2019learning, lee2020learning}. Rather than learning from a snapshot of the current robot's state, they incorporate the history of the robot's proprioceptive state \cite{DBLP:journals/corr/abs-2201-08117}. These approaches show the potential to extrapolate its learning to unseen related scenarios.

\ph{Predictive slip detection (PrePARE)} In this work, we propose a learning-based proprioceptive slip-predictive model to detect potential slip and fall events in advance. We adopt a proactive strategy 
to prepare the robot for seamless traversability in unstructured and perceptually challenging terrains. We leverage only the time-series sequences of the proprioceptive information of the robot to detect the failure events without using any camera or lidar-based sensing for the slip and fall prevention strategy. Thus, at each time instant, the model input is characterized by a vector rather than a (height x weight) 2D image or a 3D lidar point cloud which contains around 10,000 points in a single capture frame. We do extensive field testing of our strategy across different terrains. The contribution and highlights of this work are:
\begin{enumerate}[align=left,leftmargin=*]
    \item \ph{Architecture} We propose a learning-based control architecture for real-time switching of robot locomotion modes for agile slip-aware traversability analysis across varying terrains. 
    \item \ph{Predictive model} We propose a predictive proprioceptive model to enable real-time detection of potential failure events in robot locomotion during traversability in extreme terrains. 
    \item \ph{Training pipeline} We propose a semi-supervised learning procedure for the slip-predictive model which substantially reduces expert intervention. 
    \item \ph {Real-world testing and field deployment} We extensively test the proposed method on data from physical robots across multiple challenging real-world sites to assess the robustness and consistency of our approach.
\end{enumerate}

\ph{Outline}
In section \ref{sec:problem_formulation}, we formulate the problem of slip prediction from proprioceptive signals. In section \ref{sec:method}, we describe the architecture of learning-based predictive proprioception for agile robot exploration (PrePARE) and describe our method for learning the slip-predictive module. In the following subsections, we elaborate on the methods for data acquisition, labelling, slip prediction model training and deployment for online slip prediction. In section \ref{sec:experimental_results} we present our experimental results in terms of time delay, feature selection and future prediction. In section \ref{sec:conclusions} we draw the conclusions from our experiments.

\section{Problem formulation}\label{sec:problem_formulation}
Let us denote the state of the robot at time $t$ as $\mathbf{z}_t \in \mathcal{Z}$ where state is the two-dimensional position of the robot ($\mathbf{z}_t \in \mathbb{R}^2$), 
and the sensor measurements at time $t$ as $\mathbf{x}_t \in \mathbb{R}^{d}$ which are the positions, velocities and joint efforts of the two degrees of actuation (doa) hip and single-doa knee joints of the four legs of the robot measured using position encoders and force-torque sensors (thus making $d$=36).

\ph{Slip prediction model}
The slip prediction model may be formulated as:
\begin{equation}
    p_{slip}(y_{t+n}) = J(\mathbf{x}_{t-k:t}; \theta)
\end{equation}
where $y_t \in \{0, 1\}$ is a binary random variable which indicates whether the robot slipped at time $t$, $n\ge0$ is the look-ahead from time $t$ at which the slip is predicted, $p_{slip} \in [0, 1]$ is the probability of slip, $\mathbf{x}_{t-k:t} \in \mathcal{X}$ is the proprioceptive sensor history from time $t-k$ to time $t$, $J(.)$ is the slip prediction model and $\theta$ are the parameters of the model. The slip prediction model predicts  the probability of the robot slipping after $n$ time points, given the proprioceptive sensor measurements from the last $k$ time points. Thus the model $J$ maps the space of proprioceptive history, $\mathcal{X} \in \mathbb{R}^{k\times d}$, into a probability of slip, $J:\mathcal{X} \xrightarrow{}\mathbb{R}$.

\ph{Problem}
We search for an optimal model $J^*$ with optimal parameters $\theta^*$ in the space of all models, $J \in \mathcal{J}$, and corresponding space of parameters, $\theta \in \mathcal{P}$ which minimizes the error between the predicted slips and actual slips given the proprioceptive history of the robot. 

\begin{multline}
\label{eq:slip_prediction_model}
    J^*, \theta^* = \argmax_{J, \theta} [y_{t+n}\cdot\log{J(\mathbf{x}_{t-k:t}, \theta)} + \\(1-y_{t+n})\cdot \log (1-J(\mathbf{x}_{t-k:t}, \theta))]
\end{multline}

\ph{Challenge}
Finding an optimal model from a space of all models and corresponding space of parameters is computationally intractable. In the next sections, we describe a predictive proprioceptive slip prediction model that transforms this problem into its computationally tractable version. This model can be used for real-time slip detection enabling agile slip-aware traversability of legged robots in extreme terrains. 


\section{Predictive Proprioception for Agile Robot Exploration (PrePARE)}\label{sec:method}
In this section, we briefly describe the architecture for a slip-aware traversability of the legged robots, the learning-based slip-predictive model and the semi-supervised training pipeline. 

\begin{figure}[!h]
    \centering
    \includegraphics[width=0.9\linewidth]{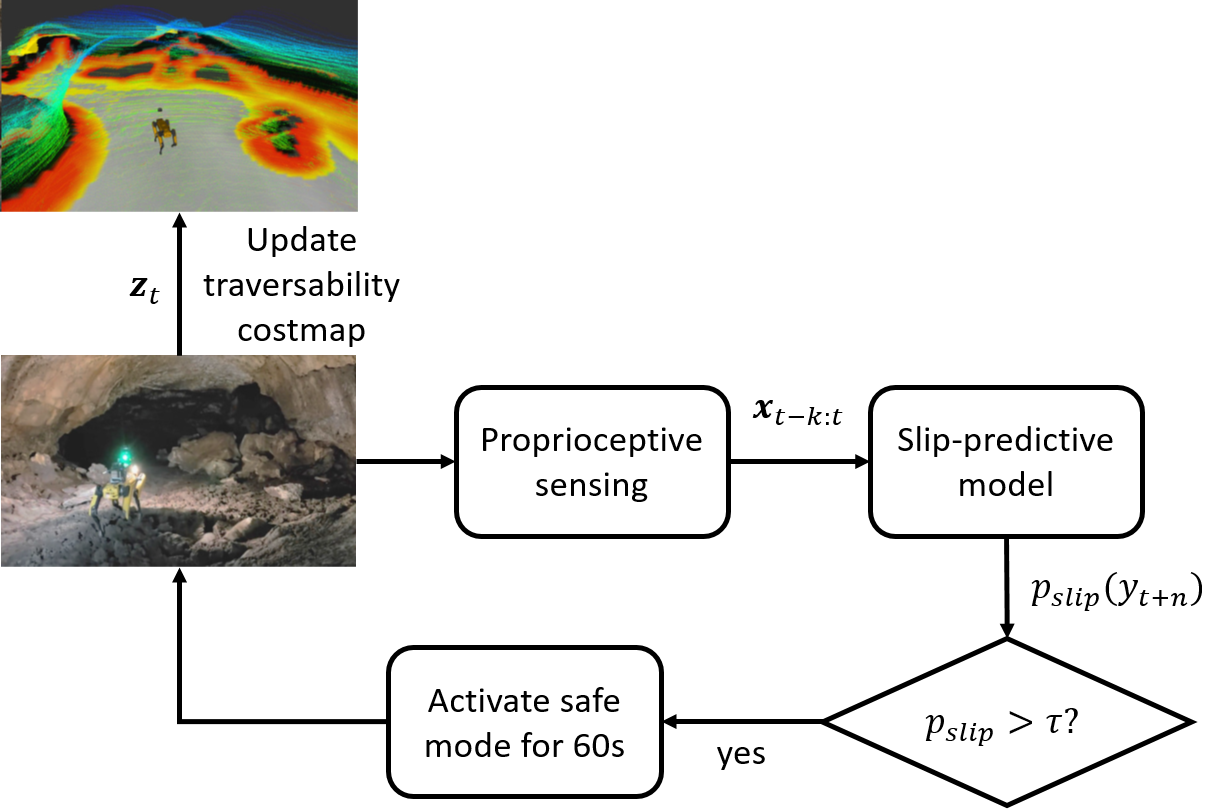}
    \caption{Overview of the slip- aware traversability architecture for legged robots. (Parts of the figure adapted from: \cite{fan2021learning}) }
    \label{fig:traversability_architecture}
\end{figure}

\ph{Architecture}
Fig. \ref{fig:traversability_architecture} shows the overview of a slip-aware traversability architecture for legged robots walking in extreme terrains. As the robot traverses through challenging terrains, the proprioceptive state of the robot is measured using proprioceptive sensors. The slip-predictive model 
predicts the probability of a slip event at a future time point ($p_{slip}(y_{t+n})$) based on the proprioceptive state history information ($\mathbf{x}_{t-k:t}$). If the probability of a slip event is greater than a threshold, $\tau$, a conservative locomotion mode such as crawl mode or amble mode is activated for the next 60 seconds (60 s) to allow better stability and fall prevention. In the crawl mode, three feet are always in contact with the ground, allowing a larger base of support, whereas in amble mode, one foot touches down at a time. The parameter $\tau$ can be adjusted to vary the allowable detection confidence. Based on our preliminary analysis, we found that a $\tau=0.5$ allows a good trade-off between the true and false positive rates of slip prediction. 
The robot can additionally use the slip event probability information and the information about its current position $\mathbf{z}_{t}$ to update the traversability cost map to increase the risk of traversing through a slippery region \cite{fan2021step}. 

\begin{figure*}[!h]
    \centering
    \includegraphics[width = 0.85\linewidth]{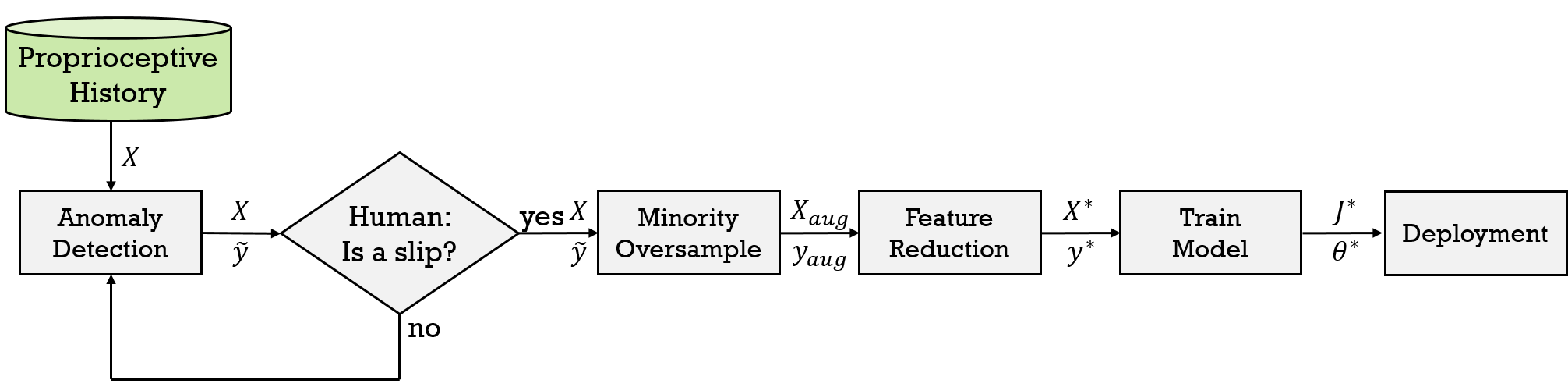}
    \caption{
    Pipeline for training the learning-based slip prediction model. First, an unsupervised anomaly detection marks anomalous sequences in the robot state history. A human annotator confirms if the event is a slip. The minority class then gets up-sampled and the features reduced. With this data an isolation forest model gets trained. For evaluation the probability of slip is predicted based on a sequence of robot states. If the probability is above a certain threshold a 'Safe Mode' gets returned.}
    \label{fig:flow_algorithm_1}
\end{figure*}

\ph{Learning-based slip-predictive model}
To make the optimization problem in Eqn. \ref{eq:slip_prediction_model} (the search for the optimal parameters of an optimal slip-predictive model) computationally tractable, we approximate the slip-predictive model using a surrogate function-approximator  characterized by an ensemble decision learner \cite{breiman2001random}. We propose a semi-supervised learning procedure (Algorithm \ref{alg:architecture} and Fig. \ref{fig:flow_algorithm_1}) for training the slip-predictive model. This procedure is adopted considering three main points: 1) to minimize  expert annotation efforts, 2) to deal with limitations of learning-based models due to slip event rarity, and 3) to keep the relevant information from the proprioceptive history and discard the irrelevant information. In the following subsections, we elaborate the methods for data acquisition and labelling, slip prediction model training and validation. 

\begin{algorithm}
\LinesNumbered
\caption{Learning-based slip-predictive model}\label{alg:architecture}
\textbf{Inputs}: 
Proprioceptive history, $\mathbf{X} \in \mathbb{R}^{N \times (k \times d)}$ \\

\begin{enumerate}
    \item $\{\Tilde{\mathbf{y}}\} = \text{AnomalyPrediction}(\mathbf{X})$
    \\
    \item $\{\Tilde{\mathbf{y}}\} = \text{HumanInTheLoop}(\Tilde{\mathbf{y}})$
    \\
    \item $\{\mathbf{X}_{aug}, \mathbf{y}_{aug}\} = \text{MinorityOversample}(\mathbf{X}, \Tilde{\mathbf{y}})$
    \\
    \item $\{\mathbf{X}^{*}, \mathbf{y}^{*}\}$ = FeatureAblation($\{\mathbf{X}_{aug}, \mathbf{y}_{aug}\}$)
    \\
    \item $J^*, \theta^* = \text{TrainModel}(\{\mathbf{X}^*, \mathbf{y}^*\})$ 
    \\
    \item Field deployment:
    \begin{enumerate}[nosep]
        \item $p_{slip}(y_{t+n}) = J^*(\mathbf{x}_{t-k:t}, \theta^*)$
        \\
        
        \item \textbf{if} {$p_{slip}(y_{t+n}) > 0.5$} \textbf{then}\\
        \quad{\textbf{return} {'Safe Mode'}}{}
                    
    \end{enumerate}

\end{enumerate}
\end{algorithm}

\ph{Data} The dataset used in this study were acquired using a 12 degrees of freedom (dof) quadruped robot (Boston Dynamics Spot) that traversed various terrains in different locations such as:
\begin{enumerate}
    \item Valentine Cave at Lava Beds National Park, CA, which is a natural old lava tube, with steeply inclined and rugged surfaces (Fig. \ref{fig:spot_diff_sites}a and Fig. \ref{fig:spot_diff_sites}f)
    \item Wells Cave in Pulaski County, KY, which is a limestone cave with plenty of rubbles, rough and narrow passages and low ceilings (Fig. \ref{fig:spot_diff_sites}b)
    \item Abandoned subway terminal in Los Angeles, CA, with dusty terrain, cluttered and narrow doorways (Fig. \ref{fig:spot_diff_sites}c and Fig. \ref{fig:spot_diff_sites}d) and with wide spaces and feature-less surfaces (Fig. \ref{fig:spot_diff_sites}e). 
    \item Mammoth Cave at Lava Beds National Park, CA, which is a lava tube with sandy, sloping floors and occasional large piles of rubble. 
\end{enumerate}
The dataset consists of joint positions, velocities and efforts from six runs of duration \textapprox 30 mins each acquired from the internal position encoders and the force-torque sensors of the Spot placed at the two-doa hip and single-doa knee joints of the robot.

\ph{Human-in-the-loop slip event annotation} We adopt a human-in-the-loop slip event annotation that consist of two stages: first, an unsupervised anomaly detection is done to mark anomalous sequences in the data. Second, the narrowed search space of the anomalous events are validated to be slip events in a replay simulation by a human annotator with knowledge of the experimental conditions.

\subsubsection*{Unsupervised anomalous sequence prediction}

We use an outlier detection algorithm, named \textit{isolation forest} \cite{liu2008isolation}, to detect anomalous occurrences in the data. The choice of the algorithm was based on our use-case situation in which slip to non-slip event ratio is highly imbalanced. 
The isolation forest is a tree-based ensemble method similar to that of the random forest \cite{breiman2001random}. To construct a path, an isolation forest randomly selects a feature at each node to split the samples using a random value of that feature. The number of nodes that a sample passes through defines the path length of that sample. The average path length across all the decision estimators of the forest is used as the measure of the normality of the sample. 

Since it is important to maximize the true positives, we apply a window of one second centered at each detected anomalous sample, so that slip events around these samples could be captured. All the frames outside these windows are considered to be normal (non-outlier). Finally, we consider all the outlier frames for further human labelling. The effort reduction for expert annotation is computed as the ratio of frames detected as normal by the anomaly detector to the total number of frames.  

\subsubsection*{Human-feedback}

By visual inspection of the robot locomotion within the outlier-detected frames in a replay simulation, we label the frames where the robot slipped. An initial inspection of the joint state data revealed that the proprioceptive signals depicted different patterns during normal locomotion and slip and thus can be efficiently used for identifying slips (Fig. \ref{fig:input_features}).

\ph{Slip-event rarity}
Imbalanced dataset is a generic problem for most learning-based algorithms \cite{krawczyk2016learning}. Our scenario is  analogous where the slip events are rare compared to the non-slip events. We adopt a minority oversampling approach to augment the slip events to match the cardinality of the non-slip events. We use a support vector-based borderline oversampling method \cite{nguyen2011borderline} for the minority oversampling. In this oversampling method, the support vectors  learned by a support vector machine (SVM) classifier from the training data are used as the borderline between the minority and majority classes. New samples belonging to the minority class are generated along the borderline (line joining support vectors of minority class) by interpolation or extrapolation based on the class identity of the nearest neighbors. 

\ph{Model training}
We use an ensemble of decision learners \cite{breiman2001random} to map the proprioceptive history into a probability of slip occurrence. The slip occurrences, $y_{t+n}$ are represented at the terminal nodes or the leafs of the decision learner, 
and a set of proprioceptive histories, $\{\mathbf{x}_{t-k:t} \in \mathbb{R}^{k\times d}\}$, leading to the slip occurrences or non-occurrences are represented at the intermediate nodes, starting from the root node. The decision learner splits recursively on an optimal value $\zeta$ of one of the features of the robot proprioceptive history, $u \in \mathbf{x}_{t-k:t}$ decided by a measure named \textit{gini impurity} \cite{suthaharan2016decision} until the terminal nodes are reached. This builds a network of decision paths leading to a tree-structured learned model. The learning overview is summarized in Algorithm \ref{alg:slip_predicitve model}. 

\begin{algorithm*}
\caption{Learning-based slip-predictive model}\label{alg:slip_predicitve model}
\textbf{Inputs}: 
Proprioceptive history, \begin{math}\mathbf{X} \in \mathbb{R}^{N \times (|t-k:t| \times d)} \end{math}, slip events,  \begin{math}\mathbf{y}\in\mathbb{R}^{N}\end{math}\\

\For {learner, $E_p,  p \in \{1\dots{P}\}$}{
    Select a bootstrap sample, $S_p \subset  \{(X,y)\} $ \\
    Let the learner parameters, $\Theta_p^* = \{\}$\\
    \While{ level, $l \leq {depth}_{max}$ }{
      \For {each node, $\nu_l$ at level $l$ }
      {Let proprioceptive history information at node, $\nu_l$ be $\mathbf{V} \in S_p$\\
      \For {candidate split $\theta = (u, \zeta), u \subset S_p$, being a feature and $\zeta$, a split value of $u$}
      {Split $\mathbf{V}$ into $\mathbf{V}^{left}(\theta) = \mathbf{V}(u<\zeta)$ and $\mathbf{V}^{right}(\theta) = \mathbf{V}(u>\zeta)$ \\
      Quality of a split is given by a loss function $Q(.)$. Let,
      \begin{math}
       G(\theta, \mathbb{V}) = \frac{|\mathbf{V}^{left}|}{|\mathbf{V}|} Q (\mathbf{V}^{left}(\theta)) + 
      \frac{|\mathbf{V}^{right}|}{|\mathbf{V}|} Q(\mathbf{V}^{right}(\theta))
      \end{math}}
      }
      Select $\theta^* = \argmin_{\theta} G(\theta, \mathbf{V})$\\
      $\Theta_p^* = \Theta_p^* \cup \theta^*$\\
      Assign $\mathbf{V}^{left}(\theta^*)$ and $\mathbf{V}^{right}(\theta^*)$ to child nodes. 
      }
     } 
    $p_{slip}(y_{t+n}) = \frac{1}{P}\sum E_p(\mathbf{x}_{t-k:t}, \Theta_p^*)$
\end{algorithm*}

\ph{Hyperparameter optimization and cross-validation} The hyper-paramters of the model (number of estimators, $P$, and the maximum depth, $d_{max}$) are optimized using a grid-search cross-validation with a grid of $P=\{50, 100, 200, 500, 1000\}$ and $d_{max} = \{5, 10, 15, 20, 25\}$. The performance of the models are evaluated using a five-fold cross validation. During each iteration of cross-validation, 80\% of the dataset is used for training and the remaining 20\% is held out for test. The train and test data consisted of same proportion of positive and negative slip samples. The train dataset is augmented using minority oversampling (see previous subsection). No oversampling is done on the test data.

\ph{Evaluation metrics} To evaluate the performance of the slip prediction model, we calculate the precision, recall and F-score of predictions (to take the class imbalance into account). 

\begin{align}
    \text{Precision} &= \frac{\text{TPR}}{\text{TPR}+\text{FPR}}\cdot 100\%\\
    \text{Recall} &= \frac{\text{TPR}}{\text{TPR}+\text{FNR}}\cdot 100\%\\
    \text{F-score} &= \frac{2*\text{Precision}*\text{Recall}}{\text{Precision}+\text{Recall}}\cdot 100\%
\end{align}
where TPR is the true positive rate, FPR, the false positive rate, and FNR, the false negative rate. 

\ph{Field deployment} Finally, we deploy our proprioceptive slip-predictive model on the Spot robot to predict slip events in real-time and autonomously switch its gait mode. If the slip probability $p_{slip}$ is above 0.5, the robot switches to a safer mode such as crawl or amble mode for the next 60 seconds. If no further slip events are detected during this window, the robot switches back to the normal mode.

\section{Experimental Results}\label{sec:experimental_results}
In this section, we present the results of our analysis on data recorded from the Spot robot walking at various challenging terrains.

\begin{figure*}[!h]
    \centering
    \includegraphics[width=\linewidth]{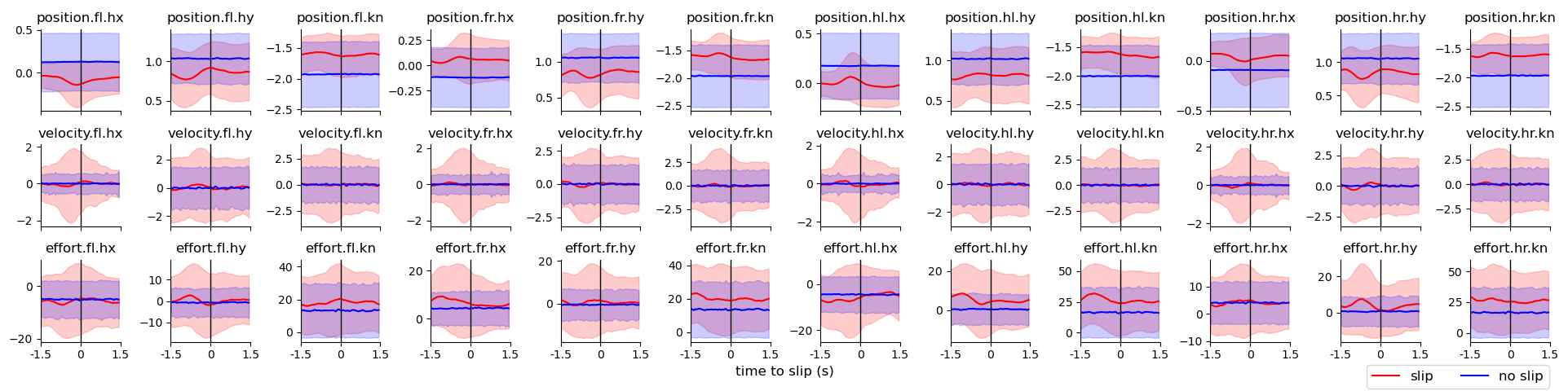}
    \caption{Mean and standard deviation of different input features for a time window of 50 samples around slip (in red). The mean and standard deviation of an equally sized window around no-slip are shown in blue (the samples around which the no-slip window is created are sampled randomly). }
    \label{fig:input_features}
\end{figure*}

\ph{Proprioceptive slip detection} First, we analyse whether proprioceptive information is useful in predicting the slips of the robot. For this we visualize the distribution of the proprioceptive signals around the slip samples and non-slip samples. We observe that although the distributions overlap, the proprioceptive data around the slip samples are quantitatively and qualitatively different from that around the non-slip samples (Fig. \ref{fig:input_features}). This prior information is leveraged to build a surrogate function-approximator for predicting the probability of occurences of slip events in advance. 

\ph{Unsupervised anomaly detection} Our unsupervised anomaly detection method is able to detect more than 98\% of all the actual slip events as anomalies. After considering a contingency window of one second around each anomalous detection, this reduced the search space for slip events for a human expert by 67\%. 

\begin{figure*}[!h]
    \centering
    \includegraphics[width=\linewidth]{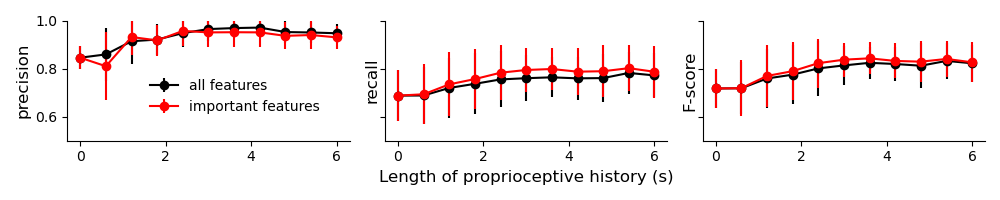}
    \caption{Evaluation of using different lengths of proprioceptive history information on the model performance. Black curves show the performance of models using complete history information and red curves show the performance of models using only the important  information in the history of same length. }
    \label{fig:time_delay_embedding}
\end{figure*}

\ph{Time-delay embedding} Next, we compute the optimal amount of proprioceptive history information required to predict the slips of the robot. For this, we train models with different amount of proprioceptive history information (from zero to six seconds) to predict the slip at the current time point. We find that increasing the length of proprioceptive history information to the model improved the model performance until a history of 3.6 s (Fig. \ref{fig:time_delay_embedding}). Using this history window, a precision of 95\%, recall of 80\%, and F-score of 84\% was obtained. This means that 95\% of all slips predicted by our models are actual slips. Additionally, our models are able to detect 80\% of all the slips. For a history greater than 
3.6 s, the model performance plateaued. Thus, we decided to use a proprioceptive history of 3.6 s for our further analyses. 

\begin{figure*}[!h]
    \centering
    \includegraphics[width=\linewidth]{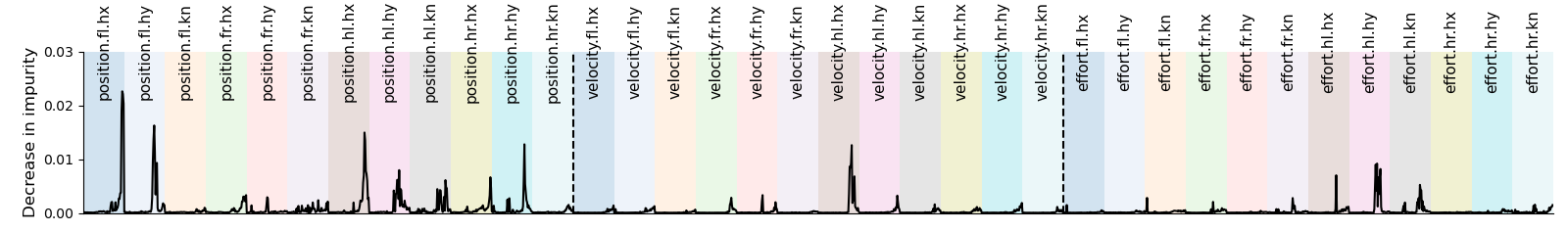}
    \caption{Importance of different input features used as inputs to the model. A history information of 3.6 seconds ($t-3.6:t$) from each proprioceptive signal (position, velocity and effort) is used as the input to the full model. }
    \label{fig:feat_imp}
\end{figure*}

\ph{Feature selection} A history window of 3.6s (60 samples) for 36 proprioceptive samples make the feature space 2160-dimensional. To reduce the feature space dimensionality, we perform a feature importance analysis using the mean decrease in impurity of each feature as the metric of importance. We find that for most of the sensor signals, information in the near history are more important than the ones further into the past. Moreover, many of the position signals are found to be important compared to velocity and effort signals (Fig. \ref{fig:feat_imp}). The input features that have very low importance are not considered further for fitting the models (for each proprioceptive signal, the time points which are found to be less important within a history of 60 samples are removed). We find that the model which did not consider these features gives comparable or better performance than that of a full model (Fig. \ref{fig:time_delay_embedding}). 

\begin{figure*}[!h]
    \centering
    \includegraphics[width=\linewidth]{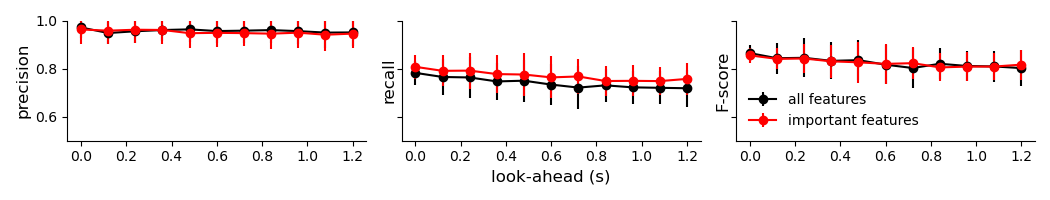}
    \caption{F-score of slip predictions into the future using a complete 3.6s history of robot state history (black) and using only the important instances of the robot state history in the 3.6s window (red). The important features are obtained using mean decrease in impurity by using this feature. }
    \label{fig:slip_forecasting}
\end{figure*}

\ph{Future prediction} Further, we train the models to predict the slip in advance (of the expert annotation) from the optimal proprioceptive history information. To analyze how much in advance the model could predict the slip satisfactorily, we train the models to predict actual slips using proprioceptive history windows of same length but with the most recent history at different time points in the past. We consider the time difference between the most recent proprioceptive history input to the model and actual slip time as the look-ahead. We find that the models can predict the slips with the best accuracy when the most recent proprioceptive history is available (look-ahead = 0, Fig. \ref{fig:slip_forecasting}). Additionally, our models can also forecast slips in advance of an expert annotated slip event using a proprioceptive history window of 3.6 s. Our analysis shows that the slip prediction accuracy drops only slightly with the increase in the look-ahead. For a look-ahead of 720 ms, a precision of 95\%, recall of 77\%, and F-score of 82\% is obtained. A prediction delay of 10 ms is observed on a PC, which leaves more than 700 ms for planning slip recovery and fall prevention maneuvers.  
The high accuracy of the look-ahead slip predictions indicates that the proprioceptive signals encode information about a potential failure event much before a human expert can detect it from visual inspection. This is also evident from Fig. \ref{fig:input_features}, which shows the difference in proprioceptive signals around expert-annotated slip events. 

\begin{figure*}[!h]
    \centering
    \includegraphics[width=0.95\linewidth]{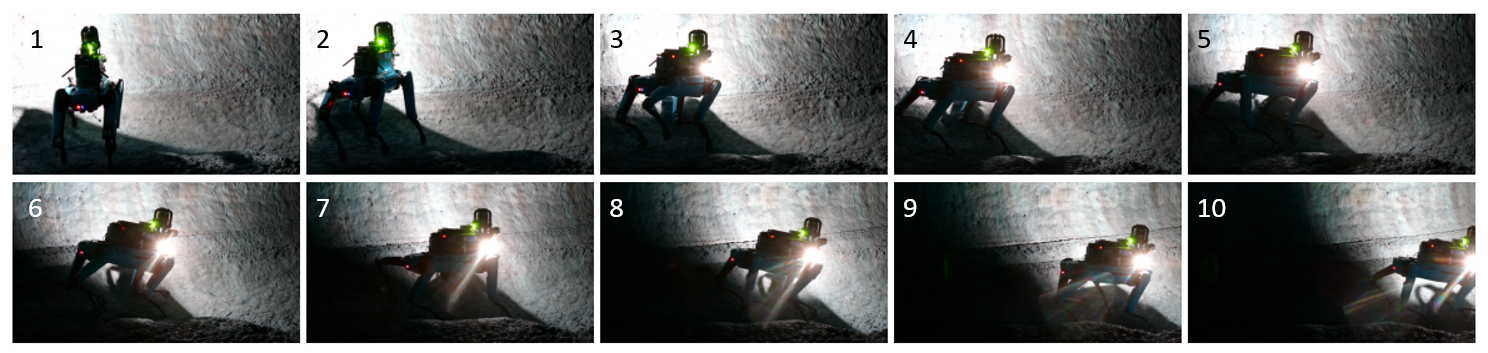}
    \caption{Time series of frames showing the Spot robot in crawl mode during field deployment near a sloped surface in a cave after detecting a slip event.}
    \label{fig:field_deployment}
\end{figure*}

\ph{Field deployment} To visually verify that the slip prediction module is reacting to an actual slip event, we used models with a look-ahead of zero for field deployment. We find that the slip prediction model successfully predicts slip events in real-time when deployed on the Spot robot. Correspondingly, the gait controller switches the robot to a safer locomotion mode. Fig. \ref{fig:field_deployment} shows such an instance where the gait controller switched the robot to crawl mode near a sloped surface. 


\section{Conclusions}\label{sec:conclusions}

We propose a predictive proprioception model to detect slip and fall events in advance from the proprioceptive history information of a quadruped robot. We designed and implemented a semi-supervised pipeline for training the predictive model for reducing expert intervention and dealing with sparsity of failure events. We test these models on data recorded from multiple sites with slippery terrains. The proposed models are able to predict slips well in advance of an expert annotator by leveraging only the proprioceptive history information. This showcases the potential of proprioception signals for developing real-time and computationally efficient failure detection modules for robots exploring extreme terrains. 
Finally, we deployed and tested the model on the Spot robot across multiple sites to assess the real-time capabilities. 
Taken together, our results suggest that predictive proprioception for failure event detection is a promising approach towards preparing robots for agile and fully autonomous exploration of extreme terrains. 


\section*{ACKNOWLEDGMENT}

We thank Tony Tran for contributing the video recordings of the field testing. 
The research was carried out at the Jet Propulsion Laboratory, California Institute of Technology, under a contract with the National Aeronautics and Space Administration (80NM0018D0004).  \copyright 2022. All rights reserved.

\bibliographystyle{ieeetr}
\bibliography{reference}

\begin{thebibliography}{10}

\bibitem{peng2020learning}
X.~B. Peng, E.~Coumans, T.~Zhang, T.-W. Lee, J.~Tan, and S.~Levine, ``Learning
  agile robotic locomotion skills by imitating animals,'' {\em arXiv preprint
  arXiv:2004.00784}, 2020.

\bibitem{melchiorri2000slip}
C.~Melchiorri, ``Slip detection and control using tactile and force sensors,''
  {\em IEEE/ASME transactions on mechatronics}, vol.~5, no.~3, pp.~235--243,
  2000.

\bibitem{Anandan2015CapacitiveTS}
N.~Anandan and B.~George, ``Capacitive tactile sensor with slip detection
  capabilities for robotic applications,'' {\em 2015 IEEE International
  Instrumentation and Measurement Technology Conference (I2MTC) Proceedings},
  pp.~464--469, 2015.

\bibitem{Teshigawara2009HighSS}
S.~Teshigawara, S.~Shimizu, K.~Tadakuma, M.~Aiguo, M.~Shimojo, and M.~Ishikawa,
  ``High sensitivity slip sensor using pressure conductive rubber,'' {\em 2009
  IEEE Sensors}, pp.~988--991, 2009.

\bibitem{Kajita2004BipedWO}
S.~Kajita, K.~Kaneko, K.~Harada, F.~Kanehiro, K.~Fujiwara, and H.~Hirukawa,
  ``Biped walking on a low friction floor,'' {\em 2004 IEEE/RSJ International
  Conference on Intelligent Robots and Systems (IROS) (IEEE Cat.
  No.04CH37566)}, vol.~4, pp.~3546--3552 vol.4, 2004.

\bibitem{focchi2018slip}
M.~Focchi, V.~Barasuol, M.~Frigerio, D.~G. Caldwell, and C.~Semini, ``Slip
  detection and recovery for quadruped robots,'' in {\em Robotics Research},
  pp.~185--199, Springer, 2018.

\bibitem{focchi2020heuristic}
M.~Focchi, R.~Orsolino, M.~Camurri, V.~Barasuol, C.~Mastalli, D.~G. Caldwell,
  and C.~Semini, ``Heuristic planning for rough terrain locomotion in presence
  of external disturbances and variable perception quality,'' in {\em Advances
  in Robotics Research: From Lab to Market}, pp.~165--209, Springer, 2020.

\bibitem{jenelten2019dynamic}
F.~Jenelten, J.~Hwangbo, F.~Tresoldi, C.~D. Bellicoso, and M.~Hutter, ``Dynamic
  locomotion on slippery ground,'' {\em IEEE Robotics and Automation Letters},
  vol.~4, no.~4, pp.~4170--4176, 2019.

\bibitem{Camurri2017ProbabilisticCE}
M.~Camurri, M.~F. Fallon, S.~Bazeille, A.~Radulescu, V.~Barasuol, D.~G.
  Caldwell, and C.~Semini, ``Probabilistic contact estimation and impact
  detection for state estimation of quadruped robots,'' {\em IEEE Robotics and
  Automation Letters}, vol.~2, pp.~1023--1030, 2017.

\bibitem{Bledt2018ContactMF}
G.~Bledt, P.~M. Wensing, S.~Ingersoll, and S.~Kim, ``Contact model fusion for
  event-based locomotion in unstructured terrains,'' {\em 2018 IEEE
  International Conference on Robotics and Automation (ICRA)}, pp.~1--8, 2018.

\bibitem{hwangbo2019learning}
J.~Hwangbo, J.~Lee, A.~Dosovitskiy, D.~Bellicoso, V.~Tsounis, V.~Koltun, and
  M.~Hutter, ``Learning agile and dynamic motor skills for legged robots,''
  {\em Science Robotics}, vol.~4, no.~26, p.~eaau5872, 2019.

\bibitem{lee2020learning}
J.~Lee, J.~Hwangbo, L.~Wellhausen, V.~Koltun, and M.~Hutter, ``Learning
  quadrupedal locomotion over challenging terrain,'' {\em Science robotics},
  vol.~5, no.~47, p.~eabc5986, 2020.

\bibitem{DBLP:journals/corr/abs-2201-08117}
T.~Miki, J.~Lee, J.~Hwangbo, L.~Wellhausen, V.~Koltun, and M.~Hutter,
  ``Learning robust perceptive locomotion for quadrupedal robots in the wild,''
  {\em CoRR}, vol.~abs/2201.08117, 2022.

\bibitem{fan2021step}
D.~D. Fan, K.~Otsu, Y.~Kubo, A.~Dixit, J.~Burdick, and A.-A. Agha-Mohammadi,
  ``Step: Stochastic traversability evaluation and planning for risk-aware
  off-road navigation,'' {\em arXiv preprint arXiv:2103.02828}, 2021.

\bibitem{fan2021learning}
D.~D. Fan, A.-A. Agha-Mohammadi, and E.~A. Theodorou, ``Learning risk-aware
  costmaps for traversability in challenging environments,'' {\em IEEE Robotics
  and Automation Letters}, vol.~7, no.~1, pp.~279--286, 2021.

\bibitem{breiman2001random}
L.~Breiman, ``Random forests,'' {\em Machine learning}, vol.~45, no.~1,
  pp.~5--32, 2001.

\bibitem{liu2008isolation}
F.~T. Liu, K.~M. Ting, and Z.-H. Zhou, ``Isolation forest,'' in {\em 2008
  eighth ieee international conference on data mining}, pp.~413--422, IEEE,
  2008.

\bibitem{krawczyk2016learning}
B.~Krawczyk, ``Learning from imbalanced data: open challenges and future
  directions,'' {\em Progress in Artificial Intelligence}, vol.~5, no.~4,
  pp.~221--232, 2016.

\bibitem{nguyen2011borderline}
H.~M. Nguyen, E.~W. Cooper, and K.~Kamei, ``Borderline over-sampling for
  imbalanced data classification,'' {\em International Journal of Knowledge
  Engineering and Soft Data Paradigms}, vol.~3, no.~1, pp.~4--21, 2011.

\bibitem{suthaharan2016decision}
S.~Suthaharan, ``Decision tree learning,'' in {\em Machine Learning Models and
  Algorithms for Big Data Classification}, pp.~237--269, Springer, 2016.

\end{thebibliography}

\clearpage

\end{document}